\def\cvprPaperID{5707}
\def\confName{ICCV}
\def\confYear{2023}

\def\paperTitle{Evaluation and Improvement of Interpretability for Self-Explainable Part-Prototype Networks}

\def\authorBlock{
    Qihan Huang\textsuperscript{$1$},
    Mengqi Xue\textsuperscript{$1$},
    Wenqi Huang\textsuperscript{$2$},
    Haofei Zhang\textsuperscript{$1$}, \\
    Jie Song\textsuperscript{$1,3,\dagger$},
    Yongcheng Jing\textsuperscript{$4$},
    Mingli Song\textsuperscript{$1$}\\
    \textsuperscript{$1$}Zhejiang University,
    \textsuperscript{$2$}Digital Grid Research Institute, China Southern Power Grid,\\
    \textsuperscript{$3$}Zhejiang University - China Southern Power Grid Joint Research Centre on AI,\\
    \textsuperscript{$4$}The University of Sydney\\
    
    {\tt\small \{qh.huang,mqxue,haofeizhang,sjie,brooksong\}@zju.edu.cn},\\
    {\tt\small huangwqcsg@163.com, yjin9495@uni.sydney.edu.au}
}

\newif\ifreview 
\newif\ifarxiv 
\newif\ifcamera \newcommand{\cameraready}{\cameratrue}
\newif\ifrebuttal

\cameraready

\pdfoutput=1
\documentclass[10pt,twocolumn,letterpaper]{article}
\ifreview \usepackage[review]{cvpr} \fi
\ifarxiv \usepackage[pagenumbers]{cvpr} \fi
\ifrebuttal \usepackage[rebuttal]{cvpr} \fi
\ifcamera \usepackage{cvpr} \fi

\usepackage{graphicx}
\usepackage{amsmath}
\usepackage{amssymb}
\usepackage{booktabs}

\usepackage{times}
\usepackage{microtype}
\usepackage{epsfig}
\usepackage[table,xcdraw]{xcolor}
\usepackage{caption}
\usepackage{float}
\usepackage{placeins}
\usepackage{color, colortbl}
\usepackage{stfloats}
\usepackage{enumitem}
\usepackage{tabularx}
\usepackage{xstring}
\usepackage{multirow}
\usepackage{xspace}
\usepackage{url}
\usepackage{subcaption}
\usepackage{xcolor}
\usepackage[hang,flushmargin]{footmisc}
\usepackage{bbm}
\usepackage{graphicx}
\usepackage{amsthm}
\usepackage{makecell}
\usepackage{bbding}
\usepackage{amsmath}
\definecolor{mygreen}{RGB}{52, 157, 2}

\usepackage{paralist}
\usepackage[accsupp]{axessibility}
\ifcamera \usepackage[accsupp]{axessibility} \fi

% ...

\ifarxiv  \fi

\newcommand{\R}[1]{{%
    \textbf{%
        \ifstrequal{#1}{1}{\textcolor{red}{R#1}}{%
        \ifstrequal{#1}{2}{\textcolor{blue}{R#1}}{%
        \ifstrequal{#1}{3}{\textcolor{magenta}{R#1}}{%
        \ifstrequal{#1}{4}{\textcolor{teal}{R#1}}{%
                           \textcolor{cyan}{R#1}%
        }}}}%
    }%
}}

\usepackage{xr-hyper}

\makeatletter
\newcommand*{\addFileDependency}[1]{
  \typeout{(#1)}
  \@addtofilelist{#1}
  \IfFileExists{#1}{}{\typeout{No file #1.}}
}

\makeatother

\usepackage[pagebackref,breaklinks,colorlinks]{hyperref}
\usepackage[capitalize]{cleveref}
\crefname{section}{Sec.}{Secs.}
\crefname{table}{Tab.}{Tabs.}
\crefname{figure}{Fig.}{Figs.}

\begin{document}
\title{\paperTitle}
\author{\authorBlock}
\maketitle
\footnotetext[1]{$\dagger$ Corresponding author.}%

\begin{abstract}
Part-prototype networks (e.g., ProtoPNet, ProtoTree, and ProtoPool) have attracted broad research interest for their intrinsic interpretability and comparable accuracy to non-interpretable counterparts.
However, recent works find that the interpretability from prototypes is fragile, due to the semantic gap between the similarities in the feature space and that in the input space.
In this work, we 
strive to address
this challenge by making the first attempt to quantitatively and objectively evaluate the interpretability of the part-prototype networks.
Specifically, we propose two evaluation metrics, termed as ``consistency score'' and ``stability score'', to evaluate the explanation consistency across images and the explanation robustness against perturbations, respectively, both of which are essential for explanations taken into practice.
Furthermore, we propose an elaborated part-prototype network with a shallow-deep feature alignment~(SDFA) module and a score aggregation~(SA) module to improve the interpretability of prototypes.
We conduct systematical evaluation experiments and provide substantial discussions to uncover the interpretability of existing part-prototype networks.
Experiments on three benchmarks across nine architectures demonstrate that our model achieves significantly superior performance to the state of the art, in both the accuracy and interpretability.
Our code is available at \textit{~\url{https://github.com/hqhQAQ/EvalProtoPNet}}.
\end{abstract}
\section{Introduction}
\label{sec:intro}


Part-prototype networks are recently emerged deep self-explainable models for image classification, which achieve excellent performance in an interpretable decision-making manner.
In particular, ProtoPNet~\cite{chen19ProtoPNet} is the first part-prototype network, with the follow-up part-prototype networks~(\eg, ProtoTree~\cite{meike2021ProtoTree}, ProtoPool~\cite{Ryma2021Assignment}, TesNet~\cite{jiaqi2021TesNet}, and ProtoPShare~\cite{Dawid2021ProtoPShare}) built upon its framework.
At its core, part-prototype networks define multiple trainable prototypes that represent specific object parts and emulate human perception by comparing object parts across images to make predictions.
Currently, part-prototype networks have been extended to various domains~(\eg, graph neural network~\cite{zhang2022ProtGNN}, deep reinforcement learning~\cite{kenny2023reinforcement}, and image segmentation~\cite{sacha2023ProtoSeg}).

\begin{figure}[t]
\centering
    \includegraphics[width=\linewidth]{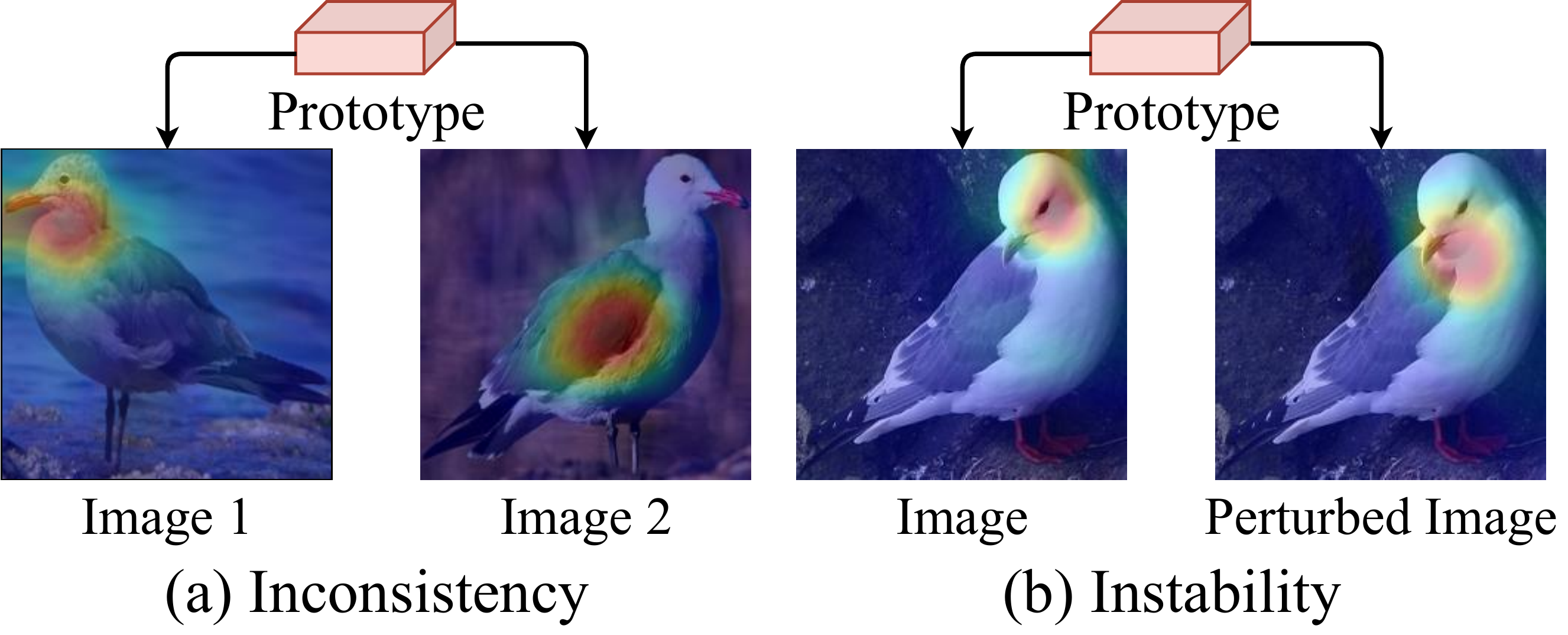}
\caption{(a) \textbf{Inconsistency.} A prototype may mistakenly correspond to different object parts in different images. (b) \textbf{Instability.} A prototype may mistakenly correspond to different object parts in the original image and the slightly perturbed image. The samples are from ProtoPNet trained on ResNet34 backbone~\cite{he2016resnet}.}
\label{fig:intro_example1}
\end{figure}

However, the current part-prototype networks only demonstrate their interpretability with only a few visualization examples, and ignore the problem that the learned prototypes of part-prototype networks do not have adequately credible interpretability~\cite{hoffmann2021doeslook, kim2022hive, Zohaib2022Transparency}.
The reasons for such unreliable interpretability are twofold: \textbf{(1) Inconsistency.} The basic design principle of part-prototype networks~\cite{chen19ProtoPNet} is that \textit{each prototype is associated with a specific object part}, but it is not guaranteed that the corresponding object part of a prototype is consistent across images, as shown in \cref{fig:intro_example1}~(a);
\textbf{(2) Instability.} Previous interpretability methods~\cite{David2018Robust, Yeh2019infidelity, zhou2021evaluating} claim that the explanation results should be stable, but the prototype in part-prototype networks is easily mapped to a vastly different object part in a perturbed image~\cite{hoffmann2021doeslook}, as shown in \cref{fig:intro_example1}~(b).
Recently, Kim \etal~\cite{kim2022hive} have proposed a human-centered method named HIVE to evaluate the interpretability of part-prototype networks. Nevertheless, HIVE requires redundant human interactions and the evaluation results are subjective.
Therefore, for the further research on the part-prototype networks, there is an urgent need for more formal and rigorous evaluation metrics that can quantitatively and objectively evaluate their interpretability.

In this work, we strive to take one further step towards the interpretability of part-prototype networks, by making the first attempt to quantitatively and objectively evaluate the interpretability of part-prototype networks, rather than the qualitative evaluations by several visualization examples or subjective evaluations from humans. 
To this end, we propose two evaluation metrics named ``consistency score'' and ``stability score'', corresponding to the above inconsistency and instability issues.
Specifically, the consistency score evaluates whether and to what extent a learned prototype is mapped to the same object part across different images.
Meanwhile, the stability score measures the robustness of the learned prototypes being mapped to the same object part if the input images are slightly perturbed.
In addition, our evaluation metrics generate objective and reproducible evaluation results using object part annotations in the dataset.
With the proposed metrics, we make the first systematic quantitative evaluations of existing part-prototype networks in \cref{sec:benchmark}. Experiments demonstrate that current part-prototype networks are, in fact, not sufficiently interpretable.

\begin{figure}[t]
\centering
    \includegraphics[width=\linewidth]{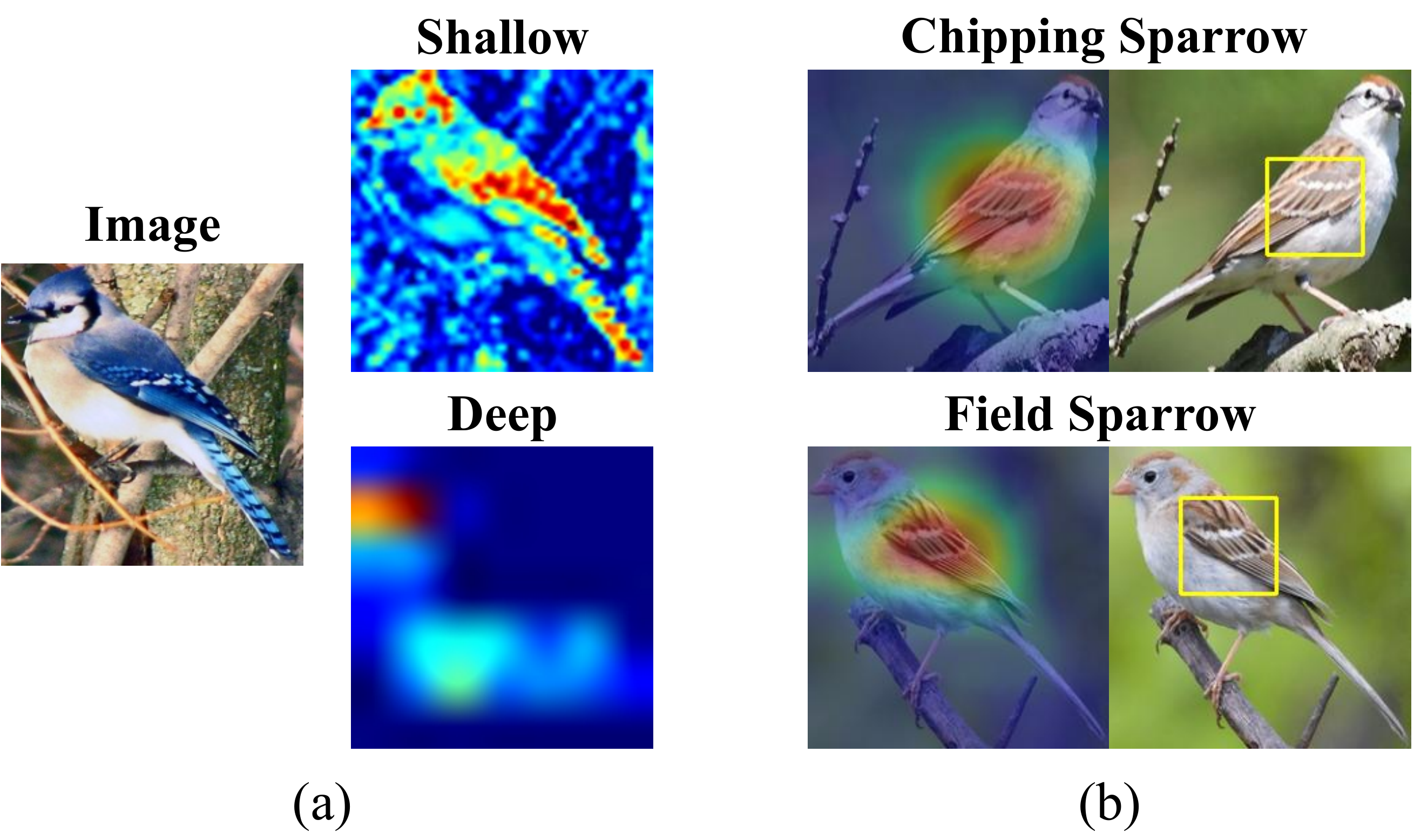}
\caption{(a) Shallow feature maps of an image retain spatial information that deep feature maps lack (the features are from one channel of the feature map). (b) A prototype from Chipping Sparrow tends to match the wing, but meanwhile it has to paradoxically ignore almost the same wing of Field Sparrow. The samples are from ProtoPNet trained on ResNet34 backbone.}
\label{fig:intro_example2}
\end{figure}

To strengthen the interpretability of prototypes, we propose an elaborated part-prototype network built upon a revised ProtoPNet with two proposed modules: a shallow-deep feature alignment~(SDFA) module and a score aggregation~(SA) module.
These two modules aim to accurately match the prototypes with their corresponding object parts across images, benefiting both consistency and stability scores.
Part-prototype networks match prototypes with object parts in two steps: (1) feature extraction of object parts; (2) matching between prototypes and features of object parts.
SDFA module improves the first step by promoting deep feature maps to spatially align with the input images. Specifically, SDFA module aligns the spatial similarity structure of shallow and deep feature maps with the observation that shallow feature maps retain spatial information that deep feature maps lack~(\cref{fig:intro_example2}~(a)).
Meanwhile, SA module improves the second step based on the observation that the matching of each prototype with its corresponding object part is disturbed by other categories~(\cref{fig:intro_example2}~(b)).
To mitigate this problem, SA module aggregates the activation values of prototypes only into their allocated categories to concentrate the matching between prototypes and their corresponding object parts.

We perform extensive experiments to validate the performance of our proposed model. Experiment results demonstrate that without using any object part annotations in training, our model achieves the state-of-the-art performance in both interpretability and accuracy on CUB-200-211 dataset~\cite{wah2011caltech}, Stanford Cars dataset~\cite{krause2013dogs} and PartImageNet dataset~\cite{he2022PartImageNet}, over six CNN backbones and three ViT backbones. Furthermore, experiment results show that the proposed consistency and stability scores are strongly positively correlated with accuracy in part-prototype networks, nicely reconciling the conflict between interpretability and accuracy in most prior interpretability methods.

To sum up, the key contributions of this work can be summarized as follows:

\begin{compactitem}
    \item We establish a benchmark to quantitatively evaluate the interpretability of prototypes of part-prototype networks with the proposed evaluation metrics (consistency score and stability score), uncovering pros and cons of various part-prototype networks.
    \item We propose an elaborated part-prototype network built upon ProtoPNet with a shallow-deep feature alignment~(SDFA) module and a score aggregation~(SA) module to enhance its interpretability.
    \item Experiment results verify that our proposed model significantly outperforms existing part-prototype networks by a large margin, in both accuracy and interpretability. Besides, the consistency and stability scores are positively correlated with accuracy, nicely reconciling the conflict between interpretability and accuracy.
\end{compactitem}
\section{Related Work}
\label{sec:related}

\subsection{Part-Prototype Networks}

ProtoPNet~\cite{chen19ProtoPNet} is the first work of part-prototype networks which define interpretable prototypes to represent specific object parts for image classification.
ProtoPNet has explicit explanations of DNNs and comparable performance with its analogous non-interpretable counterpart, which inspires many variants of ProtoPNet.
ProtoTree~\cite{meike2021ProtoTree} aggregates prototype learning into a decision tree, which generates local explanations of prototypes through a specific route of the decision tree.
TesNet~\cite{jiaqi2021TesNet} organizes the prototypes on the Grassmann manifold with several regularization loss functions as constraints.
Deformable ProtoPNet~\cite{Jon2022Deform} proposes deformable prototypes, which consist of multiple prototypical parts with changeable relative positions to capture pose variations.
PW-Net~\cite{kenny2023reinforcement} extends part-prototype networks into deep reinforcement learning and ProtoPDebug~\cite{bontempelli2022concept} proposes to utilize part-prototypes to correct the mistakes of network.
However, these methods rely on the assumption that deep features of networks retain spatial information which is not guaranteed, and they lack a quantitative metric to evaluate their explanation results.

\subsection{Evaluation of Interpretability Methods}

\begin{figure*}[t]
\centering
    \includegraphics[width=\linewidth]{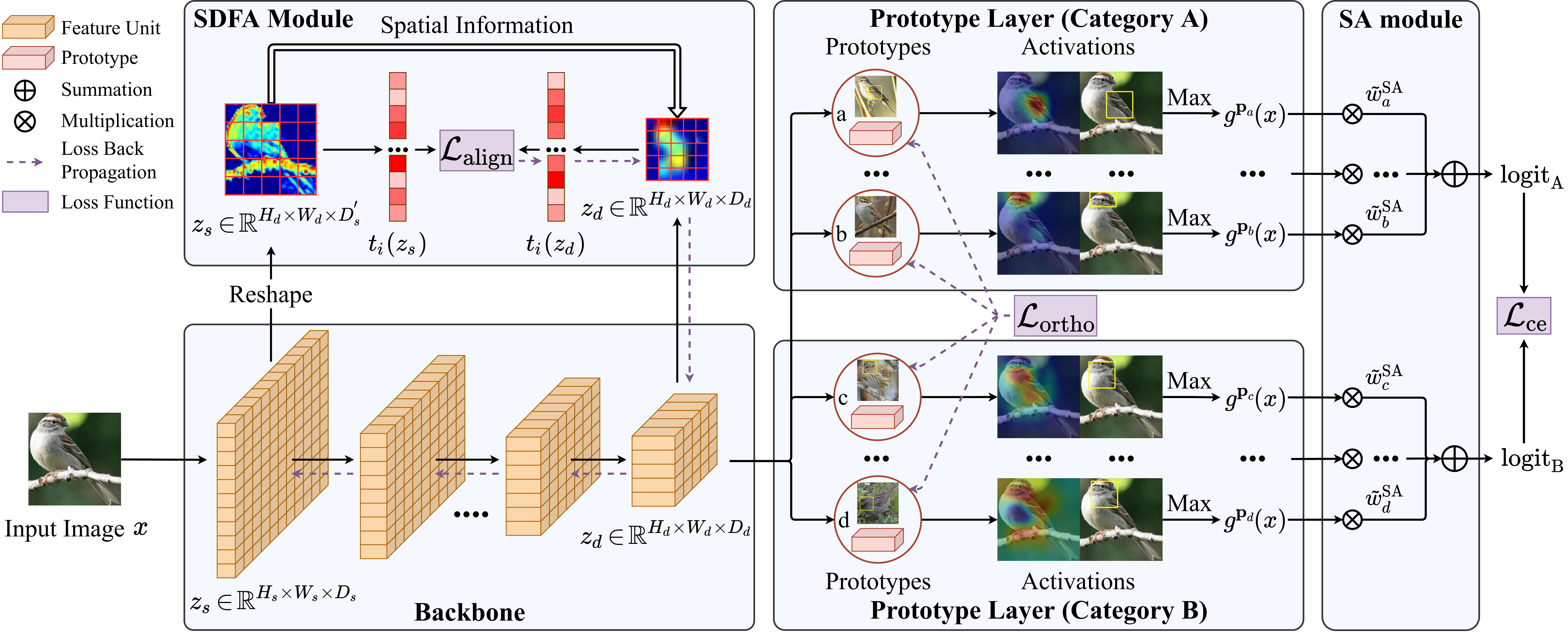}
\caption{Overview of our proposed model (only two categories are presented for brevity). Backbone is a deep convolutional network to extract the features of input image $x$. SDFA module incorporates spatial information from shallow layers into deep layers by aligning the spatial similarity structure in deep layers with that in shallow layers using an alignment loss $\mathcal{L}_{\mathrm{align}}$. The last feature map $z_d$ is fed into prototype layer for different categories. In the prototype layer, each prototype ${\boldsymbol{\mathrm{p}}}_i$ generates an activation map on $z_d$ and selects the maximum value as the activation value $g^{{\boldsymbol{\mathrm{p}}}_i}(x)$. Finally, SA module aggregates the activation values of prototypes into their allocated categories for classification. Note that $D_s^{'} = \frac{H_s}{H_d}\!\cdot\! \frac{W_s}{W_d}\! \cdot\!D_s$, $\mathcal{L}_{\mathrm{clst}}$, $\mathcal{L}_{\mathrm{sep}}$ and the loss back propagation of $\mathcal{L}_{\mathrm{ce}}$ are omitted for brevity in the figure.}
\label{fig:method}
\end{figure*}

With the development of many tasks in computer vision~(\eg, image classification~\cite{he2016resnet,su2021bcnet,su2021prioritized,su2022vitas}, detection \cite{zhai2022exploring,zhai2022one,xi2018detection,xi2022data}, and generation~\cite{jing2020dynamic,yang2023diffusion,jing2022learning}, 3D object processing~\cite{feng2023exploring,jing2021amalgamating,jing2023deep}, dataset condensation~\cite{liu2023slimmable,yu2023dataset,liu2022dataset}, and model reassembly~\cite{yang2022deep,yang2022factorizing}), numerous XAI~(\ie, explainable AI) methods and the corresponding evaluation methods are also proposed.
Zhou \etal~\cite{zhou2021evaluating} propose that current interpretability evaluation methods can be categorized into human-centered methods~\cite{Bansal2019beyond, lage2019human, chang2009reading} and functionality-grounded methods~\cite{Guidotti2019Black, Molnar2020Interpretable, slack2019assessing, Nguyen2020quantitative, Yeh2019infidelity, Mukund2017Axiomatic, Hooker2019ROAR, Montavon2018Methods, Ancona2018gradient}.
Human-centered methods require end-users to evaluate the explanations of XAI methods, which typically demand high labor costs and cannot guarantee reproducibility.
Conversely, functionality-grounded methods utilize the formal definition of XAI methods as a policy to evaluate them.
Model size, runtime operation counts assess the quality of explanations according to their explicitness, \eg, a shallow decision tree tends to have better interpretability.
Many functionality-grounded methods are also proposed to evaluate the eminent attribution-based XAI methods.
However, these functionality-grounded methods are not directed against part-prototype networks, and our work aims to propose quantitative metrics to evaluate the interpretability of part-prototype networks.

\section{Method}
\label{sec:method}

\subsection{Preliminaries}

Existing part-prototype networks are built upon the framework of ProtoPNet~\cite{chen19ProtoPNet}, and this section specifies this framework.
ProtoPNet mainly consists of a regular convolutional network $f$, a prototype layer $g_{\boldsymbol{\mathrm{p}}}$ and a fully-connected layer $h$ ($w^h$ denotes the parameters of $h$).
The prototype layer $g_{\boldsymbol{\mathrm{p}}}$ contains $M$ learnable prototypes $\boldsymbol{\mathrm{P}} = \{ \boldsymbol{\mathrm{p}}_j \in \mathbb{R}^{1 \times 1 \times D} \}_{j=1}^{M}$ for total $K$ categories ($D$ is the dimension of prototypes).

Given an input image $x$, ProtoPNet uses the convolutional network $f$ to extract the feature map $z = f(x)$ of $x$ ($z \in \mathbb{R}^{H \times W \times D}$).
The prototype layer $g_{\boldsymbol{\mathrm{p}}}$ generates an activation map $v^{{\boldsymbol{\mathrm{p}}}_j}(x) \in \mathbb{R}^{H \times W}$ of each prototype $\boldsymbol{\mathrm{p}}_j$ on the feature map $z$ by calculating and concatenating the similarity score between $\boldsymbol{\mathrm{p}}_j$ and all units $\tilde{z} \in \mathbb{R}^{1 \times 1 \times D}$ of $z$ ($z$ consists of $H \times W$ units).
Then the activation value $g^{\boldsymbol{\mathrm{p}}_j}(x)$ of $\boldsymbol{\mathrm{p}}_j$ on $x$ is computed as the maximum value in $v^{{\boldsymbol{\mathrm{p}}}_j}(x)$:

\begin{equation}
\begin{aligned}
    g^{\boldsymbol{\mathrm{p}}_j}(x) &= \max \ v^{{\boldsymbol{\mathrm{p}}}_j}(x) \\ &  = \max_{\tilde{z} \in \mathrm{units}(z)} \mathrm{Sim}(\tilde{z}, \boldsymbol{\mathrm{p}}_j).
\end{aligned}
\end{equation}

Here, $\mathrm{Sim}(\cdot, \cdot)$ denotes the similarity score between two vectors~(named activation function).
ProtoPNet allocates $N$ pre-determined prototypes to each category $k$ (note that $M = N \cdot K$), and $\boldsymbol{\mathrm{P}}_k \subseteq \boldsymbol{\mathrm{P}}$ denotes the prototypes from category $k$.
Finally, the total $M$ activation values $\{ g^{\boldsymbol{\mathrm{p}}_j}(x) \}_{j=1}^{M}$ are concatenated and multiplied with the weight matrix $w^h \in \mathbb{R}^{K \times M}$ in the fully-connected layer $h$ to generate the classification logits of $x$.
Specifically, ProtoPNet sets $w^h_{k,j}$ to be positive for all $j$ with $\boldsymbol{\mathrm{p}}_j \in \boldsymbol{\mathrm{P}}_k$, and $w^h_{k,j}$ to be negative for all $j$ with $\boldsymbol{\mathrm{p}}_j \notin \boldsymbol{\mathrm{P}}_k$. This design ensures that the high activation value of a prototype increases the probability that the image belongs to its allocated category and decreases the probability that the image belongs to other categories.

After training, the value of $g^{\boldsymbol{\mathrm{p}}_j}(x)$ reflects whether the object part represented by prototype $\boldsymbol{\mathrm{p}}_j$ exists in image $x$.
Besides, ProtoPNet visualizes the corresponding region of prototype ${\boldsymbol{\mathrm{p}}}_j$ on $x$ by resizing the activation map $v^{{\boldsymbol{\mathrm{p}}}_j}(x)$ to be a visualization heatmap with the same shape as image $x$.


\subsection{Interpretability Benchmark of Part-Prototype Networks}

The interpretability benchmark of part-prototype networks includes two evaluation metrics: consistency score~(or named \textit{part-consistency score}) and stability score~(or named \textit{stability score under perturbations}).
To calculate these two metrics, we first calculate the corresponding object part of each prototype on the images.
Next, we determine the consistency of each prototype according to whether the corresponding object parts of it are consistent across different images, and determine the stability of each prototype according to whether the corresponding object parts of it are the same on the original and perturbed images.

\subsubsection{Corresponding Object Part of Prototype}

\begin{figure}[t]
\centering
    \includegraphics[width=\linewidth]{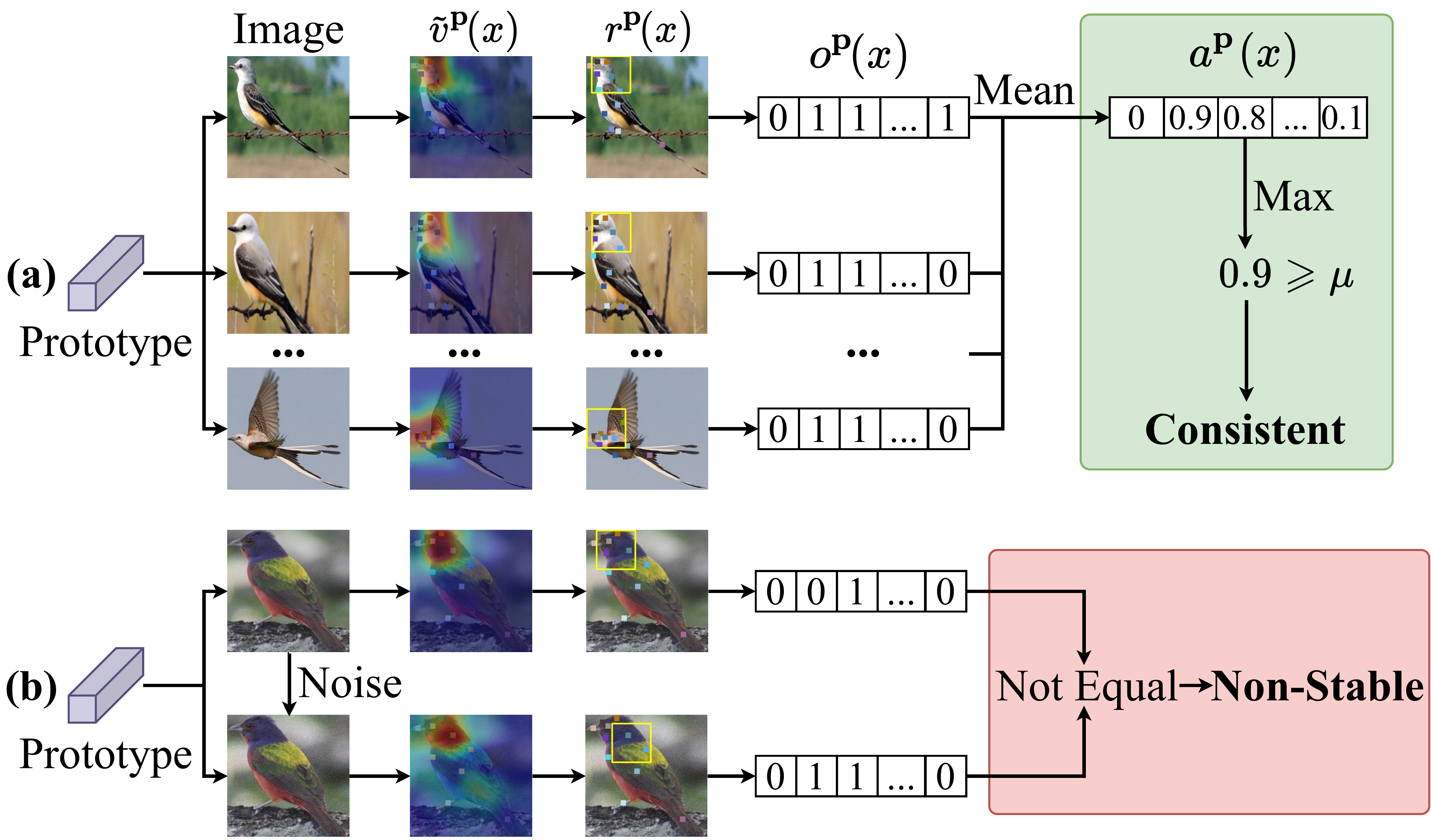}
\caption{(a) Example for determination of \textbf{consistency} of a prototype. (b) Example for determination of \textbf{stability} of a prototype. The colorful points represent the object part annotations.}
\label{fig:metric_cal}
\end{figure}

The corresponding object part of each prototype on the image is calculated using object part annotations in the dataset, which guarantees objective and reproducible evaluation results.
First, given a prototype ${\boldsymbol{\mathrm{p}}}_j$ and an input image $x$, we follow ProtoPNet to resize the activation map $v^{{\boldsymbol{\mathrm{p}}}_j}(x) \in \mathbb{R}^{H \times W}$ to be $\tilde{v}^{{\boldsymbol{\mathrm{p}}}_j}(x)$ with the same shape as $x$, then calculate the corresponding region $r^{{\boldsymbol{\mathrm{p}}}_j}(x)$ of ${\boldsymbol{\mathrm{p}}}_j$ on $x$ as a fix-sized bounding box (with a pre-determined shape $H_b \times W_b$) whose center is the maximum unit in $\tilde{v}^{{\boldsymbol{\mathrm{p}}}_j}(x)$.
Next, let $C$ denote the number of categories of object parts in the dataset, and the corresponding object part $\mathrm{o}^{\boldsymbol{\mathrm{p}}_j}(x) \in \mathbb{R}^{C}$ of prototype $\boldsymbol{\mathrm{p}}_j$ on $x$ is calculated from $r^{{\boldsymbol{\mathrm{p}}}_j}(x)$ and the object part annotations in $x$.
Specifically, $\mathrm{o}^{\boldsymbol{\mathrm{p}}_j}(x)$ is a binary vector with $\mathrm{o}^{\boldsymbol{\mathrm{p}}_j}_{i}(x) = 1$ if the $i$-th object part is inside $r^{{\boldsymbol{\mathrm{p}}}_j}(x)$ and $\mathrm{o}^{\boldsymbol{\mathrm{p}}_j}_{i}(x) = 0$ otherwise, as shown in \cref{fig:metric_cal}.
Note that ``$i$-th'' is the index of this object part in the dataset.

\subsubsection{Consistency Score}

We generate the averaged corresponding object part $a^{\boldsymbol{\mathrm{p}}_j} \in \mathbb{R}^{C}$ of each prototype $\boldsymbol{\mathrm{p}}_j$ over the 
test images from the allocated category of $\boldsymbol{\mathrm{p}}_j$, and determine the consistency of $\boldsymbol{\mathrm{p}}_j$ according to whether the maximum element in $a^{\boldsymbol{\mathrm{p}}_j}$ exceeds a threshold.
Specifically, let $\mathcal{I}_k$ denote the test images belonging to category $k$, and for each prototype $\boldsymbol{\mathrm{p}}_j$, $c(j)$ denotes the allocated category of $\boldsymbol{\mathrm{p}}_j$, and the averaged corresponding object part $a^{\boldsymbol{\mathrm{p}}_j}$ of $\boldsymbol{\mathrm{p}}_j$ is calculated as below~($\| \cdot \|$ denotes cardinality of a set):

\begin{equation}
    a^{\boldsymbol{\mathrm{p}}_j} = \frac{\sum\limits_{x \in \mathcal{I}_{c(j)}} \mathrm{o}^{\boldsymbol{\mathrm{p}}_j}(x)}{\| \mathcal{I}_{c(j)} \|}.
 \label{equa:mean_proto}
\end{equation}



For $\forall i \in \{ 1, 2, ... , C \}$, $a^{\boldsymbol{\mathrm{p}}_j}_{i} \in [0,1]$ since each $o^{\boldsymbol{\mathrm{p}}_j}_{i}$ is either 0 or 1.
If there exists an element in $a^{\boldsymbol{\mathrm{p}}_j}$ not less than a pre-defined threshold $\mu$, the prototype $\boldsymbol{\mathrm{p}}_j$ is determined to be consistent.
Finally, the consistency score $S_{\mathrm{con}}$ of a part-prototype network is defined concisely as the ratio of consistent prototypes over all $M$ prototypes~($\mathbbm{1}\{ \cdot \}$ is the indicator function):

\begin{equation}
    S_{\mathrm{con}} = \frac{1}{M}\sum\limits_{j=1}^{M}{\mathbbm{1}\{ \max(a^{\boldsymbol{\mathrm{p}}_j}) \geqslant \mu} \}.
\end{equation}

\subsubsection{Stability Score}

Stability score estimates whether prototypes retain the same corresponding object parts in the images perturbed by noise $\xi$.
As consistency score, the stability of a prototype is estimated using all the test images from its allocated category.
Finally, the stability score $S_{\mathrm{sta}}$ of a part-prototype network is calculated averagely over all prototypes:

\begin{equation}
    S_{\mathrm{sta}} \!=\! \frac{1}{M} \! \sum\limits_{j=1}^{M} \! \frac{\sum_{x \in \mathcal{I}_{c(j)}} \mathbbm{1}\{ \mathrm{o}^{\boldsymbol{\mathrm{p}}_j}(x) = \mathrm{o}^{\boldsymbol{\mathrm{p}}_j}(x + \xi) \}}{\| \mathcal{I}_{c(j)} \|}.
\label{equa:stability_score}
\end{equation}

We implement two types of noise $\xi$: (1) \textbf{Random noise.} In this way, the noise $\xi$ is randomly sampled from the same Gaussian Distribution for all models: $\xi \sim \mathcal{N}(0, \sigma^{2})$.
(2) \textbf{Adversarial noise.} We utilize the famous PGD-attack method~\cite{madry2018pgd} to generate adversarial noise, which attempts to perturb the activation map of prototypes.

\subsection{Towards a Stronger Part-Prototype Network}

The interpretability benchmark for part-prototype networks can be established using the proposed consistency and stability scores.
Next, we propose an elaborated part-prototype network built upon a revised ProtoPNet with two proposed modules: shallow-deep feature alignment~(SDFA) module and score aggregation~(SA) module.
Part-prototype networks match prototypes with object parts in two steps: (1) feature extraction of object parts; (2) matching between prototypes and features of object parts.
SDFA and SA modules respectively optimize these two steps to concentrate the matching between prototypes and their corresponding object parts, improving both consistency and stability scores.

\subsubsection{Shallow-Deep Feature Alignment Module}

The shallow-deep feature alignment~(SDFA) module is proposed to improve the feature extraction of object parts. Part-prototype networks extract the features of object part as the feature unit $\tilde{z}$ with the corresponding spatial position in $z$, requiring that deep feature maps preserve spatial information and spatially align with the input images.
However, this requirement is not guaranteed and leads to inaccurate feature extraction of object parts.
According to previous work~\cite{luo2016understanding, araujo2019computing} and our pre-experiments, units of shallow feature maps have small effective receptive fields and thereby retain spatial information.
Therefore, SDFA module preserves the spatial information of deep feature maps by incorporating spatial information from shallow layers into deep layers.

To this end, SDFA module utilizes the spatial similarity structure to represent the spatial information of a feature map and constrain the feature map of deep layers to have the identical spatial similarity structure with that of shallow layers, which is inspired by Kornblith \etal~\cite{simon2019cka} that similarity structures within representations can be used to compare two representations.
Specifically, the spatial similarity structure $t(z) \in \mathbb{R}^{HW \times HW}$ of a feature map $z \in \mathbb{R}^{HW \times D}$ ($z$ is resized from $H \times W \times D$ to $HW \times D$ for convenience) is defined as a matrix whose element represents the similarity between two units in $z$:

\begin{equation}
    t_{i,j}(z) = \mathrm{Sim}(z_{i}, z_{j}).
\end{equation}

We adopt cosine similarity for $\mathrm{Sim}(\cdot, \cdot)$ here, because cosine similarity is invariant to the norm of units which differ a lot in different layers. $z_{s} \in \mathbb{R}^{H_s \times W_s \times D_s}$, $z_{d} \in \mathbb{R}^{H_d \times W_d \times D_d}$ are used to denote the feature map of a shallow layer and a deep layer, respectively.
To keep consistent with $z_{d}$, $z_{s}$ is first resized to be $H_{d}\!\times W_{d}\!\times\!(\frac{H_{s}}{H_{d}}\!\cdot\!\frac{W_{s}}{W_{d}}\!\cdot\! D_{s})$, which implies that each unit of $z_{d}$ corresponds to an ``image patch'' in $z_{s}$.
Besides, SDFA module adopts a ReLU function to restrain only the extremely dissimilar pairs, which is to stabilize model training.
Finally, the shallow-deep feature alignment loss $\mathcal{L}_{\mathrm{align}}$ is calculated as below ($\tilde{Z} = H_{d} W_{d}$):

\begin{equation}
    \mathcal{L}_{\mathrm{align}} = \frac{1}{\tilde{Z}^{2}} \sum\limits_{i=0}^{\tilde{Z} - 1}\sum\limits_{j=0}^{\tilde{Z} - 1} \max(|t_{i,j}(z_{d}) - t_{i,j}(z_{s})| - \gamma, 0).
\end{equation}

Here, $\gamma$ denotes the threshold for ReLU function. Besides, $t(z_{s})$ is set to be detached so that it never requires gradient.

\subsubsection{Score Aggregation Module}

The score aggregation~(SA) module is proposed to address the problem that the matching of each prototype with its corresponding object part is disturbed by other categories.
This problem stems from the fully-connected layer $h$ that the classification score of a category is dependent on prototypes of other categories.
The vanilla ProtoPNet has proposed a convex optimization step to mitigate this problem, which optimizes the weight $w^{h}$ of last layer by minimizing this loss: $\mathcal{L}_{\mathrm{ce}} + \sum_{k=1}^{K}\sum_{j:\boldsymbol{\mathrm{p}}_j \notin \boldsymbol{\mathrm{P}}_k} |w^h_{k,j}|$ ($\mathcal{L}_{\mathrm{ce}}$ is the classification loss).
However, we find that this optimization step will make some $w^h_{k,j}$ with $\boldsymbol{\mathrm{p}}_j \in \boldsymbol{\mathrm{P}}_k$ be negative, which causes high activation values of these prototypes to paradoxically contribute negatively to their allocated categories and hurts the interpretability of the model.

Instead, our work directly addresses this problem by replacing the fully-connected layer with the SA module.
SA module aggregates the activation values of prototypes only into their allocated categories, followed by a learnable layer with weights $w^{\mathrm{SA}} \in \mathbb{R}^{M}$ to adjust the importance of all $M$ prototypes.
Specifically, let $\tilde{w}^{\mathrm{SA}}_{j} = e^{w_{j}^{\mathrm{SA}}} / (\sum_{\boldsymbol{\mathrm{p}}_i \in \boldsymbol{\mathrm{P}}_k} e^{w_{i}^{\mathrm{SA}}})$, the classification score $\mathrm{logit}_{k}$ of category $k$ is calculated in SA module as below:

\begin{equation}
    \mathrm{logit}_{k} =  \sum\limits_{\boldsymbol{\mathrm{p}}_j \in \boldsymbol{\mathrm{P}}_k} \tilde{w}^{\mathrm{SA}}_{j} \cdot g_{\boldsymbol{\mathrm{p}}_j}(x).
\end{equation}

\subsubsection{Revised Baseline \& Loss Function \label{subsec:revised_baseline}}

The vanilla ProtoPNet adopts some components for model training: a cluster loss $\mathcal{L}_{\mathrm{clst}}$, a separation loss $\mathcal{L}_{\mathrm{sep}}$. The details of them are presented in Section {\color{red} A.4} of the appendix.
Besides, we additionally adopt several simple but effective modifications on the vanilla ProtoPNet:
(1) \textbf{Activation function.} We select inner product as the activation function following TesNet~\cite{jiaqi2021TesNet} (another part-prototype network), which enlarges the gap between high activations and low activations, and thus better discriminates regions with different activations.
(2) \textbf{Orthogonality loss.} We use the orthogonality loss following Deformable ProtoPNet~\cite{Jon2022Deform} to diversify prototypes within the same category, as shown in \cref{equa:ortho_loss} ($\boldsymbol{\mathrm{P}}^k \in \mathbb{R}^{N \times D}$ denotes the concatenation of prototypes from category $k$ and $\mathbb{I}_{N}$ is an $N \times N$ identity matrix).
(3) \textbf{Hyper-parameters.} We select some different hyper-parameters, as shown in Section {\color{red} A.5} of the appendix.

\begin{equation}
\label{equa:ortho_loss}
    \mathcal{L}_{\mathrm{ortho}} = \sum\limits_{k=1}^{K} \|\boldsymbol{\mathrm{P}}^k ({\boldsymbol{\mathrm{P}}^k})^{\top} - \mathbb{I}_{N} \|^{2}.
\end{equation}

This revised ProtoPNet significantly improves the performance of the vanilla ProtoPNet, and we add SDFA and SA modules into it for further improvement~(\cref{fig:method}).
The total loss $\mathcal{L}_{\mathrm{total}}$ of our final model is as below ($\mathcal{L}_{\mathrm{ce}}$ denotes the cross entropy loss, $\lambda_{\mathrm{align}}$ denotes the coefficient of $\mathcal{L}_{\mathrm{align}}$):

\begin{equation}
    \mathcal{L}_{\mathrm{total}} = \mathcal{L}_{\mathrm{ce}} + \underbrace{\mathcal{L}_{\mathrm{clst}} + \mathcal{L}_{\mathrm{sep}} + \mathcal{L}_{\mathrm{ortho}}}_{\mathrm{Previous \ Methods}} + \underbrace{\lambda_{\mathrm{align}} \mathcal{L}_{\mathrm{align}}}_{\mathrm{Ours}}.
\end{equation}

\section{Experiments}

\subsection{Experimental Settings}

\begin{table*}
\small
\renewcommand\arraystretch{1}
\centering
\setlength{\tabcolsep}{1.4mm}{
\begin{tabular}{c|*3{c}|*3{c}|*3{c}|*3{c}|*3{c}}
  \toprule
\multirow{2}*{\small \textbf{Method}} & \multicolumn{3}{c|}{\small \textbf{ResNet34}}  & \multicolumn{3}{c|}{\small \textbf{ResNet152}} & \multicolumn{3}{c|}{\small \textbf{VGG19}} & \multicolumn{3}{c|}{\small \textbf{Dense121}} & \multicolumn{3}{c}{\small \textbf{Dense161}} \\

\cline{2-16}

& \textbf{\small Con.} & \textbf{\small Sta.} & \textbf{\small Acc.} & \textbf{\small Con.} & \textbf{\small Sta.} & \textbf{\small Acc.} & \textbf{\small Con.} & \textbf{Sta.} & \textbf{\small Acc.} & \textbf{\small Con.} & \textbf{\small Sta.} & \textbf{\small Acc.} & \textbf{\small Con.} & \textbf{\small Sta.} & \textbf{\small Acc.} \\

\midrule

{\small Baseline} & N/A & N/A & 82.3 & N/A & N/A & 81.5 & N/A & N/A & 75.1 & N/A & N/A & 80.5 & N/A & N/A & 82.2 \\
{\small ProtoTree}~\cite{meike2021ProtoTree} & 10.0 & 21.6 & 70.1 & 16.4 & 23.2 & 71.2 & 17.6 & 19.8 & 68.7 & 21.5 & 24.4 & 73.2 & 18.8 & 28.9 & 72.4 \\
{\small ProtoPNet}~\cite{chen19ProtoPNet} & 15.1 & 53.8 & 79.2 & 28.3 & 56.7 & 78.0 & 31.6 & 60.4 & 78.0 & 24.9 & 58.9 & 80.2 & 21.2 & 58.2 & 80.1 \\
{\small ProtoPool}~\cite{Ryma2021Assignment} & 32.4 & 57.6 & 80.3 & 35.7 & 58.4 & 81.5 & 36.2 & 62.7 & 78.4 & 48.5 & 55.3 & 81.5 & 40.6 & 61.2 & 82.0 \\
{\small Deformable}~\cite{Jon2022Deform} & 39.9 & 57.0 & 81.1 & 44.2 & 53.5 & 82.0 & 40.6 & 61.5 & 77.9 & 61.4 & 64.7 & 82.6 & 46.7 & 63.9 & 83.3 \\
{\small TesNet}~\cite{jiaqi2021TesNet} & 53.3 & 65.4 & 82.8 & 48.6 & 60.0 & 82.7 & 46.8 & 58.2 & 81.4 & 63.1 & 66.1 & 84.8 & 62.2 & 67.5 & 84.6 \\
\hline

{\small \bf Ours} & 52.9 & 66.3 & 82.3 & 50.7 & 65.7 & 83.8 & 44.6 & 56.9 & 80.6 & 52.9 & 59.1 & 83.4 & 56.3 & 61.5 & 84.7 \\

{\small \bf Ours + SA} & 67.5 & 69.9 & 83.6 & 59.2 & 68.4 & 84.3 & 50.4 & 60.5 & 82.1 & 65.3 & 61.5 & 84.8 & 70.0 & 66.6 & 85.8 \\

{\small \bf Ours + SA + SDFA} & \textbf{70.6} & \textbf{72.1} & \textbf{84.0} & \textbf{62.1} & \textbf{70.8} & \textbf{85.1} & \textbf{56.5} & \textbf{63.5} & \textbf{82.5} & \textbf{68.1} & \textbf{67.6} & \textbf{85.4} & \textbf{72.0} & \textbf{71.8} & \textbf{86.5} \\

\bottomrule
\end{tabular}}
\caption{The comprehensive evaluation of interpretability and accuracy of part-prototype networks on CUB-200-2011 dataset. The results are over five convolutional backbones pre-trained on ImageNet. Con., Sta. and Acc. denote consistency score, stability score and accuracy, respectively. Our results are averaged over 4 runs with different seeds. Bold font denotes the best result.}
\label{tab:cub_benchmark}

\end{table*}
\begin{table}
\renewcommand\arraystretch{1.1}
\small
\centering
\setlength{\tabcolsep}{1.0mm}{
\begin{tabular}{c|c|*3{c}}
  \toprule
\textbf{\small Method} & \textbf{\small Backbone} & \textbf{\small Con.} & \textbf{\small Sta.} & \textbf{\small Acc.} \\

\midrule

{\small ProtoTree} & ResNet50 (iN) & 16.4 & 18.4 & 82.2 \\
{\small ProtoPool} & ResNet50 (iN) & 34.6 & 45.8 & 85.5  \\
{\small Deformable} & ResNet50 (iN) & 39.7 & 48.5 & 85.6  \\
{\small Ours} & ResNet50 (iN) & \textbf{56.9} & \textbf{67.8} & \textbf{87.1}  \\

\hline

{\small ProtoPNet} & VGG + ResNet + Dense & 23.9 & 57.7 & 84.8 \\
{\small TesNet} & VGG + ResNet + Dense & 54.4 & 63.2 & 86.2 \\
{\small ProtoTree} & ResNet50 (iN) × 3 & 17.2 & 19.8 & 86.6 \\
{\small ProtoPool} & ResNet50 (iN) × 3 & 34.3 & 46.2 & 87.5 \\
{\small Ours} & ResNet50 (iN) × 3 & \textbf{56.7} & \textbf{67.4} & \textbf{88.3}  \\

\hline

{\small ProtoTree} & ResNet50 (iN) × 5 & 16.8 & 18.5 & 87.2 \\
{\small ProtoPool} & ResNet50 (iN) × 5 & 34.4 & 45.7 & 87.6 \\
{\small Ours} & ResNet50 (iN) × 5 & \textbf{57.0} & \textbf{67.5} & \textbf{88.5}  \\

  \bottomrule
  \end{tabular}}
\caption{Results of the benchmark on CUB-200-2011 dataset over ResNet50 backbone pre-trained on iNaturalist2017 dataset and combination of multiple backbones. ``× 3'' denotes combining the classification logits of 3 models trained with different seeds. Bold font denotes the best result.}
\label{tab:cub_benchmark_inat}

\end{table}

\noindent \textbf{Datasets.} We follow existing part-prototype networks to conduct experiments on CUB-200-2011~\cite{wah2011caltech} and Stanford Cars~\cite{krause2013dogs}.
CUB-200-2011 contains location annotations of object parts for each image, including 15 categories of object parts (back, breast, eye, leg, ...) that cover the bird's whole body. Therefore, our interpretability benchmark is mainly established based on this dataset.
Besides, we also validate the performance of our model on PartImageNet~\cite{he2022PartImageNet}.


\noindent \textbf{Benchmark Setup.} Our benchmark evaluates five current part-prototype networks: ProtoPNet~\cite{chen19ProtoPNet}, ProtoTree~\cite{meike2021ProtoTree}, TesNet~\cite{jiaqi2021TesNet}, Deformable ProtoPNet~\cite{Jon2022Deform} and ProtoPool~\cite{Ryma2021Assignment}.
The consistency score and stability score of these methods are re-implemented faithfully following their released codes.

\noindent \textbf{Parameters.} We set $H_b$, $W_b$, $\mu$ and $\sigma$ to be 72, 72, 0.8 and 0.2 for the interpretability evaluation of all part-prototype networks.
Our models are trained for 12 epochs with Adam optimizer~\cite{Kingma2015adam} (including 5 epochs for warm-up).
We set the learning rates of the backbone, the add-on module and the prototypes to be $1e^{-4}$, $3e^{-3}$ and $3e^{-3}$ for our method.
0.5 and 0.1 are chosen for $\lambda_{\mathrm{align}}$ and $\gamma$.
The number of prototypes per category and the dimension of prototypes are 10 and 64 for our method.
More details of our experiment setup are presented in Section {\color{red} A} of the appendix.

\subsection{Benchmark of Part-Prototype Networks \label{sec:benchmark}}

\begin{figure}[t]
\centering
    \includegraphics[width=\linewidth]{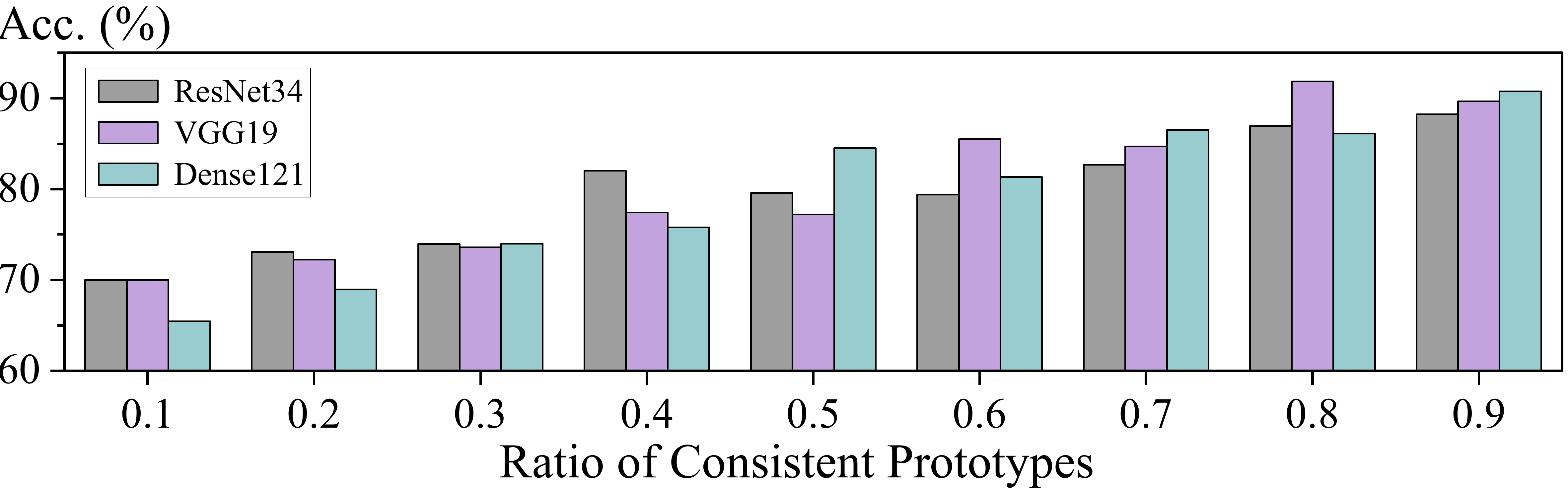}
\caption{The test accuracy on images from each class correlates positively with its ratio of consistent prototypes in ProtoPNet (over three backbones).}
\label{fig:consistency_to_acc}
\end{figure}
\begin{figure}[t]
\centering
    \includegraphics[width=\linewidth]{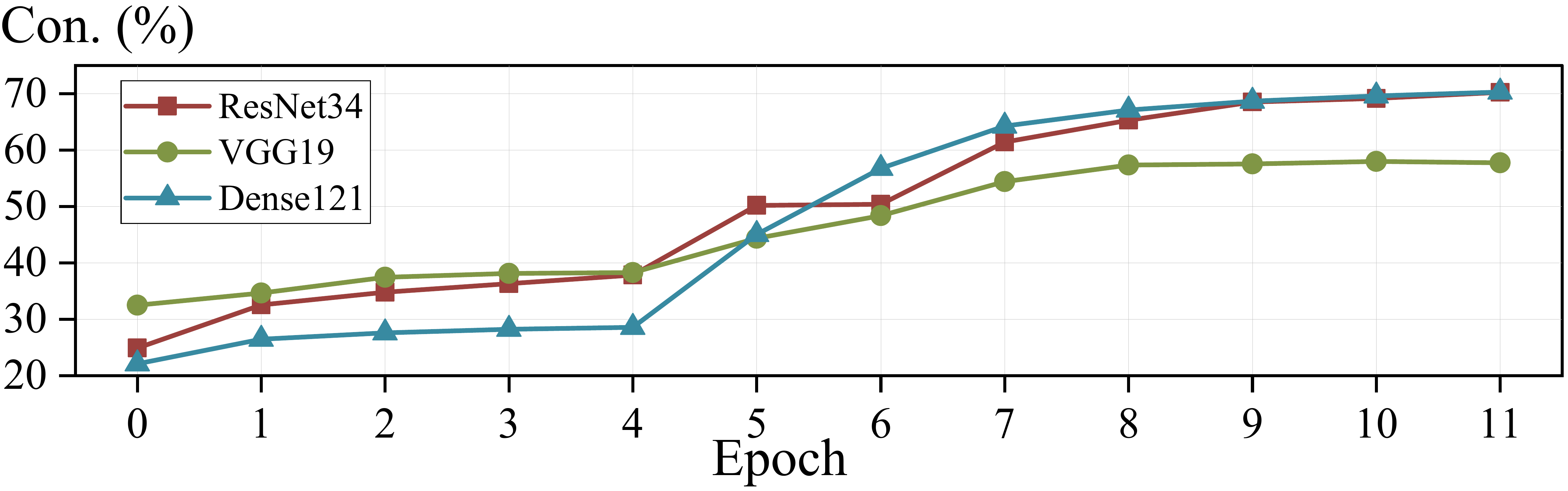}
\caption{The consistency score increases along with the training process of ProtoPNet (over three backbones).}
\label{fig:epoch_to_consistency}
\end{figure}

With the proposed consistency and stability scores, the benchmark of part-prototype networks on CUB-200-2011 can be established (over five convolutional backbones pre-trained on ImageNet), as shown in \cref{tab:cub_benchmark}.
The stability score in this table is calculated with random noise, we provide the evaluation results of other variants of stability score~(including adversarial noise) in Section {\color{red} B.3} of the appendix and find that they are consistent with the random-noise version.
In the table, ``Baseline'' is the simplest non-interpretable model with a fully-connected layer on the last feature map for classification.
Existing part-prototype networks are listed in ascending order of their accuracy in the table, and some important conclusions can be drawn from it.

\noindent\textbf{The vanilla ProtoPNet has poor interpretability.} The consistency score of the vanilla ProtoPNet ranges from 15.1 to 31.6 on five backbones, meaning that most of its prototypes are not interpretable because they cannot represent the same object parts in different images.
This points out that qualitative analysis (cherry picks) of explanation results is not reliable, and quantitative analysis is more meaningful and essential.
However, many current methods directly transfer the paradigm of the vanilla ProtoPNet to other domains (\eg, image segmentation, person re-identification, deep reinforcement learning) with only qualitative analysis.

\noindent\textbf{The accuracy of part-prototype networks correlates positively with their consistency and stability scores overall.}
For each backbone in \cref{tab:cub_benchmark}, the part-prototype network with higher accuracy generally has higher consistency and stability scores.
This phenomenon accords with the definition of part-prototype networks that a prototype represents a specific object part, and part-prototype networks make predictions by comparing the object parts which are activated by the same prototypes in the test image and training images.
In this definition, the mismatch of object parts in the test image and training images will severely drop the accuracy of the model. For example, a non-consistent and non-stable part-prototype network will make wrong predictions by mistakenly comparing the head part in the test image with the stomach part in the training images.


Besides, we conduct two experiments to analyze the relation between interpretability and accuracy of ProtoPNet.
First, we calculate the accuracy of each category and get the average accuracy of categories with the same ratio of consistent prototypes in \cref{fig:consistency_to_acc}, showing that the accuracy on different categories positively correlates with the ratio of consistent prototypes.
Second, we calculate the consistency score after each training epoch in \cref{fig:epoch_to_consistency}, showing that the consistency score increases along with the model training and thus has a positive correlation with the model accuracy.

\subsection{Comparisons with State-of-the-Art Methods}

As shown in \cref{tab:cub_benchmark}, we integrated the SDFA and SA modules into our revised ProtoPNet~(``Ours'' denotes this revised ProtoPNet), and the consistency score, stability score and accuracy of final model are significantly superior to current part-prototype networks on CUB-200-2011~(the main dataset adopted by previous methods) over five backbones.
\cref{tab:cub_benchmark_inat} demonstrates the experiment results on ResNet50 backbone pre-trained on iNaturalist2017 dataset~\cite{van2018inaturalist} and combination of multiple backbones, which also verifies the significant performance of our model.
Besides, our model achieves the best performance on Stanford Cars and PartImageNet, shown in Section {\color{red} B.1} and {\color{red} B.2} of the appendix.

\begin{table}
\renewcommand\arraystretch{1.15}
\small
\centering
\setlength{\tabcolsep}{0.94mm}{
\begin{tabular}{c|*4{c}}
  \toprule
\textbf{Method} & \textbf{ResNet34} & \textbf{ResNet152} & \textbf{VGG19} & \textbf{Dense121} \\

\midrule

{w/o SDFA} & 12.4 & 22.8 & 9.8 & 15.9 \\
{w/ SDFA} & 70.8({\color{mygreen}+58.4}) & 74.3({\color{mygreen}+51.5}) & 30.5({\color{mygreen}+20.7}) & 47.1({\color{mygreen}+31.2}) \\

\bottomrule
\end{tabular}}
\caption{Similarity (\%) between spatial similarity structures of shallow layers and deep layers without/with SDFA module on the whole test set of CUB-200-2011.}
\label{tab:simi_spatial}

\end{table}
\begin{figure}[t]
\centering
    \includegraphics[width=\linewidth]{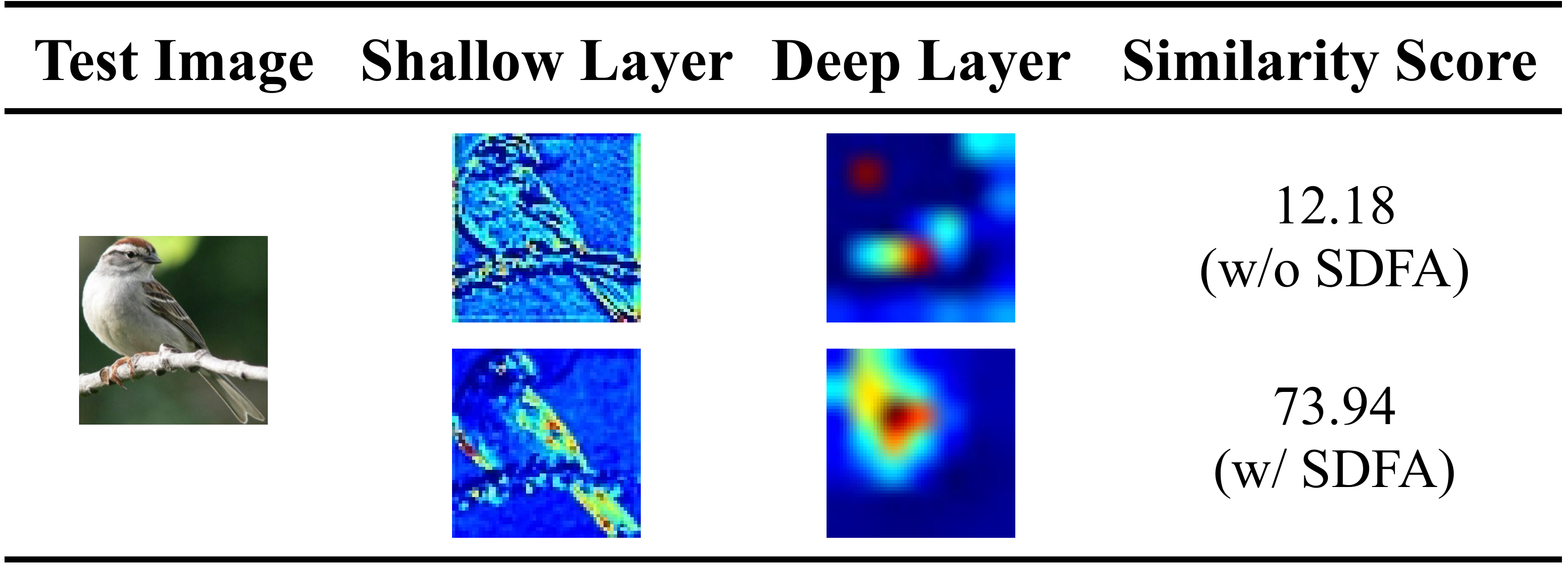}
\caption{Visualization of feature maps from shallow layers and deep layers without/with SDFA module.}
\label{fig:SDFA_vis}
\end{figure}
\begin{table}
\renewcommand\arraystretch{1.15}
\centering
\small
\setlength{\tabcolsep}{2.05mm}{
\begin{tabular}{c|*4{c}}
  \toprule
\textbf{Method} & \textbf{ResNet34} & \textbf{ResNet152} & \textbf{VGG19} & \textbf{Dense121} \\

\midrule

{w/o SA} & 3.9 & 4.7 & 7.3 & 6.2 \\
{w/ SA} & 8.5({\color{mygreen}+4.6}) & 9.8({\color{mygreen}+5.1}) & 10.8({\color{mygreen}+3.5}) & 11.8({\color{mygreen}+5.6}) \\

\bottomrule
\end{tabular}}
\caption{Average number of similar prototypes from other categories of each prototype on CUB-200-2011 dataset.}
\label{tab:simi_num_proto}

\end{table}

\subsection{Ablation Study}

\cref{tab:cub_benchmark} demonstrates that the SDFA and SA modules both effectively improve consistency score, stability score and accuracy of the model. 
We provide ablation experiments of the hyper-parameters used in our method in Section {\color{red} B.5} of the appendix.
Besides, we conduct two ablation experiments to analyze the effect of SDFA and SA modules.

\noindent\textbf{SDFA Module.} We calculate the similarity between spatial similarity structures ($t(z_{s})$ and $t(z_{d})$ with shape $\mathbb{R}^{H_{d}W_{d} \times H_{d}W_{d}}$) of shallow and deep layers with/without SDFA module, and generate the average results over all test images.
Specifically, the similarity $\mathrm{Sim}(t(z_{s}), t(z_{d}))$ between $t(z_{s})$ and $t(z_{d})$ is calculated in \cref{equa:simi_spatial} ($\tilde{Z} = H_{d} W_{d}$).
\cref{tab:simi_spatial} shows that SDFA module improves the similarity between spatial similarity structure of shallow layers and deep layers by a large margin.
Besides, \cref{fig:SDFA_vis} shows that the shallow feature maps both explicitly contain spatial information, and the deep feature map can highlight the object instead of background with SDFA module.


\begin{equation}
\setlength\abovedisplayskip{-8pt}
\setlength\belowdisplayskip{2pt}
    \mathrm{Sim}(t(z_{s}), t(z_{d})) = \frac{1}{\tilde{Z}} \sum\limits_{i=0}^{\tilde{Z}-1} e^{-\| t_{i}(z_{s}) - t_{i}(z_{d}) \|^2}.
\label{equa:simi_spatial}
\end{equation}

\noindent\textbf{SA Module.} We calculate the number of similar prototypes from other categories for each prototype and present the averaged results without/with SA module (two prototypes with cosine similarity over 0.6 are considered to be similar here).
As shown in \cref{tab:simi_num_proto}, each prototype has fewer similar prototypes from other categories without SA module, indicating that prototypes are suppressed to represent similar object parts among categories without SA module, due to the original paradoxical learning paradigm.

\subsection{Visualization Results}

\begin{figure}[t]
\centering
    \includegraphics[width=\linewidth]{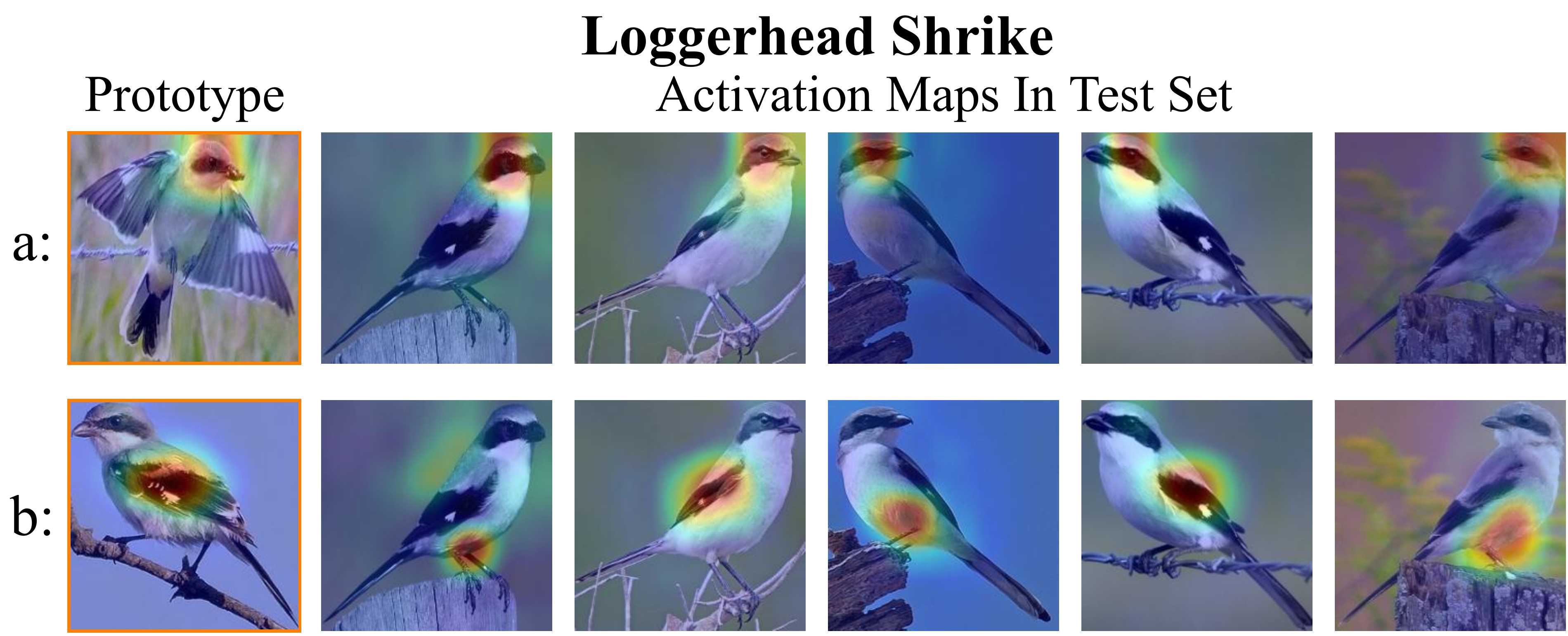}
\caption{A consistent prototype (a) and a non-consistent prototype (b). Note that $\max(a^{\boldsymbol{\mathrm{p}}_a}) = 0.97$ and $\max(a^{\boldsymbol{\mathrm{p}}_b}) = 0.33$. The first image in each row is the training image representing this prototype.}
\label{fig:consis_non_consis}
\end{figure}

\noindent\textbf{Consistency Score.} To analyze the effect of our proposed consistency score, we provide visualization of activation maps of a consistent prototype $\boldsymbol{\mathrm{p}}_a$ and a non-consistent prototype $\boldsymbol{\mathrm{p}}_b$ from Yellow Billed Cuckoo category ($\max(a^{\boldsymbol{\mathrm{p}}_a}) = 0.97$ and $\max(a^{\boldsymbol{\mathrm{p}}_b}) = 0.13$).
As shown in \cref{fig:consis_non_consis}, prototype $\boldsymbol{\mathrm{p}}_a$ consistently activates the head part in test images and training images, while prototype $\boldsymbol{\mathrm{p}}_b$ desultorily activates the wing part, belly part and feet part.


Additionally, we demonstrate more comprehensive visualization analysis on the corresponding regions of consistent prototypes from our model in Section {\color{red} C} of the appendix.



\section{Conclusion}
\label{sec:conclusion}

This work establishes an interpretability benchmark to quantitatively evaluate the interpretability of prototypes for part-prototype networks, based on two evaluation metrics (consistency score and stability score).
Furthermore, we propose a SDFA module to incorporate the spatial information from shallow layers into deep layers and a SA module to concentrate the learning of prototypes.
We add these two modules into a simply revised ProtoPNet, and it significantly surpasses the performance of existing part-prototype networks on three datasets, in both accuracy and interpretability.
Our work has great potential to facilitate more quantitative metrics to evaluate the explanation results of interpretability methods, instead of using limited visualization samples which can be easily misled by cherry picks. 
In the future, we will extend this work to other concept embedding methods towards a unified benchmark for visual concepts.

\noindent \textbf{Acknowledgements.}
This work is funded by National Natural Science Foundation of China~(61976186, U20B2066, 62106220), Ningbo Natural Science Foundation~(2021J189), and the Fundamental Research Funds for the Central Universities~(2021FZZX001-23, 226-2023-00048).

{\small
\bibliographystyle{ieee_fullname}
\bibliography{11_references}
}


\end{document}